\pdfoutput=1

\documentclass[11pt]{article}

\usepackage{emnlp2021}

\usepackage{times}
\usepackage{latexsym}
\usepackage{times}
\usepackage{latexsym}
\usepackage{array}
\usepackage{nicematrix}
\usepackage{multirow}
\usepackage{pbox}
\usepackage{tabularx}

\usepackage[T1]{fontenc}

\usepackage[utf8]{inputenc}
\usepackage{microtype}

\title{DSC-IITISM at FinCausal 2021: Combining POS tagging with Attention-based Contextual Representations for Identifying Causal Relationships in Financial Documents}

\author{Gunjan Haldar${^1}^\oplus$, \hspace{0.05cm} Aman Mittal${^1}^\oplus$, \hspace{0.05cm} Pradyumna Gupta${^2}^\oplus$\\
$^1$Department of Mechanical Engineering\\%
$^2$Department of Electronics Engineering\\%
$^\oplus$Indian Institute of Technology (ISM), Dhanbad 826004, India}

\begin{document}
\maketitle
\begin{abstract}
Causality detection draws plenty of attention in the field of Natural Language Processing and linguistics research. It has essential applications in information retrieval, event prediction, question answering, financial analysis, and market research. In this study, we explore several methods to identify and extract cause-effect pairs in financial documents using transformers. For this purpose, we propose an approach that combines POS tagging with the BIO scheme, which can be integrated with modern transformer models to address this challenge of identifying causality in a given text. Our best methodology achieves an F1-Score of 0.9551, and an Exact Match Score of 0.8777 on the blind test in the FinCausal-2021 Shared Task at the FinCausal 2021 Workshop.
\end{abstract}

\section{Introduction}

Integrating causality information as text features can substantially benefit a plethora of applications such as text mining \citep{article}, event prediction \citep{event}, question answering \citep{DBLP:journals/corr/SharpSJCH16} and many more. One of the primary motives of this, which has been explored in this challenge, is to extract causality in the financial domain, which can be applied to various tasks such as financial services support \citep{Chen2020NLPIF}, consumer review \citep{inproceedings}, stock movement prediction \citep{chen2021stock} as well as help different institutions to gain insights into the financial sector. 

By examining the financial documents carefully, one can observe that single and multiple causal events in a given paragraph may exist. Additionally, there can also be the existence of numerous causal chains in the same. To deal with such cases, we formulate causality detection and extraction task as a sequence labeling and modeling problem and propose an approach using POS tagging \citep{9074941} with BIO scheme tagging \citep{Bio} integrated with an ensemble of BERT Large-cased \citep{DBLP:journals/corr/abs-1810-04805}, XLNet Base \citep{DBLP:journals/corr/abs-1906-08237}, BERT Large-Cased Whole Word Masking, GPT-2 \citep{radford2019language} and RoBERTa Base \citep{DBLP:journals/corr/abs-1907-11692}, achieving an F1-Score of 0.9551 and Exact Match score of 0.8777 on Blind test dataset provided by the workshop.

\section{Dataset}
The dataset provided \citep{Mariko-fincausal-fnp} \citep{Mariko-fincausal-2020} for this challenge\footnote{\url{http://wp.lancs.ac.uk/cfie/fincausal2021/}} has been extracted from 2019 financial documents provided by Qwam\footnote{\url{https://www.qwamci.com/}}, consisting of the complete text and the extracted cause and effect pairs along with offset markers. It was also observed that multiple instances comprised of the same text but different causality pairs, due to presence of multiple chains of causal relationships. Total instances present in the database were 2393 which were split into 2101 training and 292 validation instances.

\section{Methodology}

\begin{table*}
\centering
\begin{tabular}{lcc}
\hline
\textbf{Token} & \textbf{POS Tag} & \textbf{BIO Tag}\\
\hline
The & \verb|DT| & \verb|B-E|\\
Sunshine & \verb|NNP| & \verb|I-E|\\ 
State & \verb|NNP| & \verb|I-E|\\ 
drew & \verb|VBD| & \verb|I-E|\\
... & ... & \verb|I-E|  \\
... & ... & \verb|I-E|  \\
older & \verb|JJR| & \verb|I-E|\\
\hline
\end{tabular}
\begin{tabular}{lcc}
\hline
\textbf{Token} & \textbf{POS Tag} & \textbf{BIO Tag}\\
\hline
It & \verb|PRP| & \verb|B-C|\\
is & \verb|VBZ| & \verb|I-C|\\
consistently & \verb|RB| & \verb|I-C|\\ 
one & \verb|CD| & \verb|I-C|\\ 
... & ... & \verb|I-C|  \\
... & ... & \verb|I-C|  \\
taxes & \verb|NNS| & \verb|I-C|  \\
\hline
\end{tabular}
\caption{Pre-processed Output stored in text format, The above text represents an example instance from the training set.}
\label{tab:Preprocessing}
\end{table*}

\subsection{Part-Of-Speech (POS) Tagging}

We tokenize each sentence and generate rule-based part-of-speech (POS) tags \citep{9074941} for each token. Rule-based POS tagging uses contextual information and a set of handwritten rules to assign POS tags to tokens in a sentence.

After tokenizing the data, the tokens are converted into POS tags. The POS tags are enumerated, which are further mapped on the tokenized sentences. These POS tags are represented in the form of a one-hot vector.  This vector is concatenated with the model’s hidden state output of the last layer, which is then sent to the final linear layer of the model. Predictions are performed on the concatenated vector or tensor.

\begin{table}[hpt]
\centering
\begin{tabular}{clc}
\hline
\textbf{Tag} & \textbf{Description} & \textbf{BIO Label}\\
\hline
\verb|B-E| & At the Beginning of Effect & 3\\
\verb|B-C| & At the Beginning of Cause & 1\\
\verb|I-C| & Inside of Cause & 2\\ 
\verb|I-E| & Inside of Effect & 4\\ 
\verb|-| & Padding & 0\\\hline
\end{tabular}
\caption{Tagging Scheme explanation. BIO tag ``O'' will be converted to padding.}
\label{tab:BIO}
\end{table}
\subsection{BIO Scheme Tagging}
To extract the causal relations and positional information of the words, considering the semantics of the causal events, we use the BIO tagging \citep{Bio} scheme i.e. Begin-Inside-Outside tagging with Cause and Effect labels (C-E). BIO tagging scheme will represent whether the token is at the beginning (B) of the target phrase, inside (I) of a target phrase and tokens which are not a part of cause or effect are considered as being outside (O) of the target phrase and  are labelled as padding (-). Additionally, due to varying sequence length, extra tokens which are not included in cause and effect tuples are converted to padding as shown in Table~\ref{tab:BIO}.

\subsection{Pre-processing}
To begin with, two different modes are given as input for pre-processing. When the mode is ``training'', the corresponding sentence and cause-effect tuples in the training data are append to a dictionary, otherwise when the mode is ``test'', sentences in the test dataset are appended to a dictionary. Each sentence in the paragraph is tokenized, subsequently, separate tokens and their positional index are stored in a list. 

Further, for the preparation of BIO tags, the index of the tokenized words are identified in each sentence using its respective index and stored in a dictionary. The beginning of cause and effect pairs are found in the sentence, and this pair is tokenized. Tokens at the beginning of the cause and effect are labelled as \verb|B-C| and \verb|B-E| respectively. Subsequent tokens in cause and effect sentences are labelled as \verb|I-C| and \verb|I-E| respectively. These labels along with the words are stored in a dictionary identified by their index. The tags are extracted from the dictionary. This process is iterated over all the instances in the training set.

To end with, each word is concatenated with its respective POS tags and BIO tags as shown in Table~\ref{tab:Preprocessing}. The pre-processed file is stored in a text format which is further passed onto the model as input.

\subsection{Transformer Architecture}

\begin{table*}[hpt]
\centering
\begin{tabular}{l |c c c c c}
\hline
\textbf{Model} & \textbf{Epochs} & \textbf{MSL${^\star}$} & \textbf{Validation Score${^\dagger}$} & \textbf{Blind Test${^\dagger}$}\\
\hline
BERT-base & 40 & 256 & 0.9197 & 0.9253\\
RoBERTa-base & 50 & 256 & 0.9201 & 0.9372\\
GPT-2 & 20  & 128 & 0.9251 & 0.9422\\
XLNet-base & 50 & 128 & 0.9368 & 0.9466\\
\textbf{BERT-large} & \textbf{50} & \textbf{256} & \textbf{0.9389} & \textbf{0.9517}\\ 
BWM & 50  & 256 & 0.9327 & 0.9476\\\hline

\end{tabular}
\caption{Model Comparison by Experimentation; $^\star$Maximum Sequence Length, \hspace{0.05cm} $^\dagger$F1 Score}
\label{tab:allExp}
\end{table*}

\begin{table*}
\centering
\begin{tabular}{l|ccccc}
\hline
\textbf{Model} & \textbf{F1-Score} & \textbf{Recall} & \textbf{Precision} & \textbf{Exact Match}\\
\hline
\rule{0pt}{12pt}BERT-large + RoBERTa + \\XLNet + GPT-2 + BWM & 0.9551  & 0.9580 & 0.9554 & 0.8777\\
\hline
\end{tabular}

\caption{Best performing method on the official Blind Test of FinCausal-2021}
\label{tab:top}
\end{table*}

For the purpose of this challenge, our best approach utilizes an ensemble developed using BERT (Bidirectional Encoder Representations from Transformers) Large-Cased model \citep{DBLP:journals/corr/abs-1810-04805}, RoBERTa (Robustly Optimized BERT Pre-training Approach) \citep{DBLP:journals/corr/abs-1907-11692}, GPT-2 (Generative Pre-trained Transformer) \citep{radford2019language}, BERT Large-Cased Whole Word Masking \citep{DBLP:journals/corr/abs-1810-04805} (BWM), XLNet \citep{DBLP:journals/corr/abs-1906-08237} by \nocite{Kao2020NTUNLPLAF} applying the huggingface\footnote{\url{https://huggingface.co/bert-large-cased}} \citep{DBLP:journals/corr/abs-1910-03771} package. 

\subsubsection{Models}
\textbf{BERT Large-cased} transformer model has been pre-trained on the English language with a masked language modeling (MLM) objective distributed into Masked Language Modelling and Next Sentence Prediction (NSP), which converges to learn an internal representation that can be utilized to extract features from downstream tasks. This model consists of 24 transformer encoder layers with 1024 hidden dimensions with 16 self-attention heads.

\textbf{BWM} model has been pre-trained on the same language corpus as BERT Large-Cased model but with a whole word masking technique, wherein all of the tokens corresponding to a word are masked at once. The overall masking rate remains the same. The model was pre-trained on 4 cloud TPUs for one million steps with a batch size of 256. The sequence length was limited to 128 tokens for 90\% of the steps and 512 for the remaining 10\%. The optimizer used is Adam with a learning rate of 1e-4, $\beta_{1}$ = 0.9 and $\beta_{2}$ = 0.999 and a weight decay of 0.01

\textbf{RoBERTa} is pre-trained with the same objective as BERT but on 1024 V100 GPUs for 500K steps with a batch size of 8K and a sequence length of 512. Adam optimizer is used with a learning rate of 6e-4, $\beta_{1}$ = 0.9, $\beta_{2}$ = 0.98 and $\epsilon$ = 1e-6, and a weight decay of 0.01 with dynamic masking where the model randomly masks 15\% of the words in the input then run the entire masked sentence through the model and has to predict the masked words.

\textbf{GPT-2} transformer model takes sequences of continuous text as input and uses an internal mask-mechanism to predict the token at any position ``i'' by the inputs at position 1 to ``i''.

\textbf{XLNet} is a generalized autoregressive pre-training method enabling learning of bidirectional contexts. 

\subsubsection{Training}
The pre-processed output file was procured, and for every instance, the corresponding POS and BIO tags of each token was extracted and stored in an array. According to the maximum sequence length, these arrays were padded. Depending on the transformer model utilized, \verb|[CLS]| and \verb|[SEP]| tokens were appended to the tokens. For instance, if the transformer model was BERT Large-Cased, \verb|[CLS]| token was appended at the beginning and \verb|[SEP]| token at the end and when the transformer model is XLNet, \verb|[CLS]| is appended at the end.  Pseudo POS tag ID and BIO tag ID for \verb|[CLS]| or \verb|[SEP]| token was set as ``0'' and ``-100'' respectively. All ID sequences were padded with padding token ID - ``0'' in POS tag sequence and ``-100'' in BIO tag sequence.

Each model consumed on an average 3-4 hours for training. The configurations of the best models which are used for the ensemble are reported in Table~\ref{tab:allExp}. All these models have been trained with a batch size of 64 with cross-entropy loss \citep{gordonrodriguez2020uses} so that only real IDs contribute to the loss function and not the padding IDs. 

\newcolumntype{L}[1]{>{\hsize=#1\hsize\raggedright\arraybackslash}X}%

\begin{table*}
\begin{tabularx}{\textwidth}{ | L{0.85} | L{0.575} | L{0.575} | }
  \hline
  \textbf{Text} & \textbf{Cause} & \textbf{Effect}\\\hline
  
  \multirow{2}{6cm}{The company also recently announced a quarterly dividend, which was paid on Tuesday, September 3rd. Shareholders of record on Thursday, August 15th were paid a \$0.03 dividend. This represents a \$0.12 annualized dividend and a yield of 3.42\%.} & The company also recently announced a quarterly dividend, which was paid on Tuesday, September 3rd. & Shareholders of record on Thursday, August 15th were paid a \$0.03 dividend. \\ 
  \cline {2-3}
   & The company also recently announced a quarterly dividend, which was paid on Tuesday, September 3rd. & This represents a \$0.12 annualized dividend and a yield of 3.42\%. \\ 
  \hline
  \multirow{2}{6cm}{If you pay the full RAD there is no interest (DAP) pay no RAD and you will pay a DAP which is the interest on the full amount: \$22,160.} & pay no RAD & you will pay a DAP which is the interest on the full amount: \$22,160. \\ 
  \cline {2-3}
   & If you pay the full RAD & there is no interest (DAP) \\ [5ex]
  \hline
\end{tabularx}
\caption{Table representing identical multi-causal chains. Causal chains in the training dataset.}
\label{tab:dupli}
\end{table*}

\subsection{Post-processing \& Exact Match Optimization}

The received predictions are in the format of tuples of tokens and their corresponding predicted BIO tag. The BIO tags are retrieved and stored in a list with the index of each token in the prediction. Further, this process is iterated over all the predicted instances and recorded. We tried to optimize the Exact Match metric by selecting the longest cause-effect pair when multiple causal chains are present in a given data instance. If the number of padding tokens was less than a given threshold between two similar predicted phrases (Cause/Effect), the two pairs were merged. 


\subsection{Ensemble}
After each prediction was extracted from different models present in the ensemble, the mode was calculated to find the most frequently occurring label. In the presence of a tie-breaker scenario, we select the label predicted by the best performing single transformer model, BERT Large-Cased. Further, after extracting all the tags, these were aligned with the text to get the actual words bundled together to form the cause-effect pair.  

\section{Experimentation and Results}

Different models along with custom loss functions were trained on the given data and local F1 score, Recall, and Precision were evaluated. Transformer models including RoBERTa (Robustly Optimized BERT Pre-training Approach) \citep{DBLP:journals/corr/abs-1907-11692}, GPT-2 (Generative Pre-trained Transformer) \citep{radford2019language}, BERT Base \citep{DBLP:journals/corr/abs-1810-04805}, BERT Large-Cased Whole Word Masking \citep{DBLP:journals/corr/abs-1810-04805} (BWM), XLNet \citep{DBLP:journals/corr/abs-1906-08237} were experimented with different hyper-parameter settings. The best performing settings along with their corresponding scores are reported in Table~\ref{tab:allExp}. The results were evaluated locally, and considering those metrics, the model performance was observed. To boost up optimization, ensembles of the aforementioned transformer models were experimented and evaluated.

GPT-2 was trained and experimented with, but due to expensive computational requirements, it was trained for 20 epochs. Loss function while RoBERTa-base transformer model was being trained on the data couldn't converge, resulting in a low metric score; similar behavior was observed in XLNet. BERT large-cased model outperformed all these models due to its large architectural layout when a single shot transformer is concerned. Maximum Sequence Length (MSL) is a critical factor while training a model with limited computational resources, because having a high MSL means most of the memory is wasted for padding and not used for weight update. Subsequently, smaller MSL values are chosen for transformer models with vast architecture. Ensembles mentioned in Table~\ref{tab:top} gave a relatively low F1 score when BERT-base was included along with other models indicating that the lower performance of BERT-base single shot experiment could be the prominent dropping factor. The performance metrics of the top approach is shown in Table~\ref{tab:top}.

\section{Conclusion}
This paper presents our sequence labeling and modeling approach, combining POS tags with BIO scheme using ensemble optimization strategy comprising BERT-large, RoBERTa, XLNet, GPT-2, and BERT-Large (whole word masking) for causality detection in financial documents which helped us achieve the highest Exact Match score of 0.8777, on the FinCausal-2021 Shared Task leaderboard.
Future works can describe an optimization pipeline constituting architecturally larger transformer models. Furthermore, more advanced post-processing strategies can be investigated to extract multiple causal relationships in a text.

\bibliography{anthology,custom}
\bibliographystyle{acl_natbib}

\end{document}